\documentclass[10pt,twocolumn,letterpaper]{article}

\usepackage{cvpr}              

\usepackage{colortbl}
\usepackage{color}
\usepackage{xcolor}
\usepackage{mathtools}
\usepackage{tikz}
\usepackage[ruled,vlined]{algorithm2e}
\newcommand{\Exp}{\mathrm{Exp}}
\newcommand{\Log}{\mathrm{Log}}
\newcommand{\Mobius}{\mathrm{M\ddot{o}bius}}

\renewcommand{\paragraph}[1]{{\vspace{0.5mm}\noindent \bf #1}}

\definecolor{cvprblue}{rgb}{0.21,0.49,0.74}
\usepackage[pagebackref,breaklinks,colorlinks,citecolor=cvprblue]{hyperref}

\def\paperID{833} 
\def\confName{CVPR}
\def\confYear{2024}

\def\etal{\textit{et al.}}
\newcommand{\Poincare}{Poincar\'{e} }
\usepackage{multirow}

\title{HyperSDFusion: Bridging Hierarchical Structures in Language and Geometry for Enhanced 3D Text2Shape Generation
}

\author{\begin{tabular}{cccccccccccccccc}\multicolumn{4}{c}{Zhiying Leng \textsuperscript{1,2}\thanks{This work was done when the author was at TUM.}} & \multicolumn{4}{c}{Tolga Birdal \textsuperscript{3}} & \multicolumn{4}{c}{Xiaohui Liang \textsuperscript{2,4}\thanks{corresponding author}} &  \multicolumn{4}{c}{Federico Tombari \textsuperscript{1}}\end{tabular}\vspace{1mm}\\
{\small \textsuperscript {1}Technical University of Munich, Germany \quad \textsuperscript{3} Imperial College London, U.K.}\\
{\small\textsuperscript{2}Beihang University, China  \quad \quad \quad \quad \quad \quad\textsuperscript{4} Zhongguancun Laboratory, China} \\
{\tt\small \{zhiyingleng,liang\_xiaohui\}@buaa.edu.cn, t.birdal@imperial.ac.uk, tombari@in.tum.de}
}

\begin{document}
\maketitle
\begin{abstract}
3D shape generation from text is a fundamental task in 3D representation learning. The text-shape pairs exhibit a hierarchical structure, where a general text like ``chair" covers all 3D shapes of the chair, while more detailed prompts refer to more specific shapes. Furthermore, both text and 3D shapes are inherently hierarchical structures. However, existing Text2Shape methods, such as SDFusion, do not exploit that. In this work, we propose HyperSDFusion, a dual-branch diffusion model that generates 3D shapes from a given text. Since hyperbolic space is suitable for handling hierarchical data, we propose to learn the hierarchical representations of text and 3D shapes in hyperbolic space. First, we introduce a hyperbolic text-image encoder to learn the sequential and multi-modal hierarchical features of text in hyperbolic space. In addition, we design a hyperbolic text-graph convolution module to learn the hierarchical features of text in hyperbolic space. In order to fully utilize these text features, we introduce a dual-branch structure to embed text features in 3D feature space. At last, to endow the generated 3D shapes with a hierarchical structure, we devise a hyperbolic hierarchical loss. Our method is the first to explore the hyperbolic hierarchical representation for text-to-shape generation. Experimental results on the existing text-to-shape paired dataset, Text2Shape, achieved state-of-the-art results. 
We release our implementation under \href{HyperSDFusion.github.io}{HyperSDFusion.github.io}.
\end{abstract}

\section{Introduction}
Text-to-Shape synthesis~\cite{text2shape2018,autosdf2022,sdfusion2023,magic3d2023,dreamfusion2022} involves the task of generating high-quality and faithful shapes given a text prompt and holds significant promise for a wide range of applications including augmented/virtual reality and design, offering the potential for automated, diverse, and cost-effective 3D content. 
Unlike image-based media~\cite{renderdiffusion2023,holodiffusion2023,sparsefusion2023}, natural language provides a more direct means of expression. However, effectively marrying the realms of 3D geometry and natural language is challenging, leading to no established standard for text-guided 3D shape generation. 

\begin{figure}[tb]
    \centering
   \includegraphics[width=1.\linewidth]{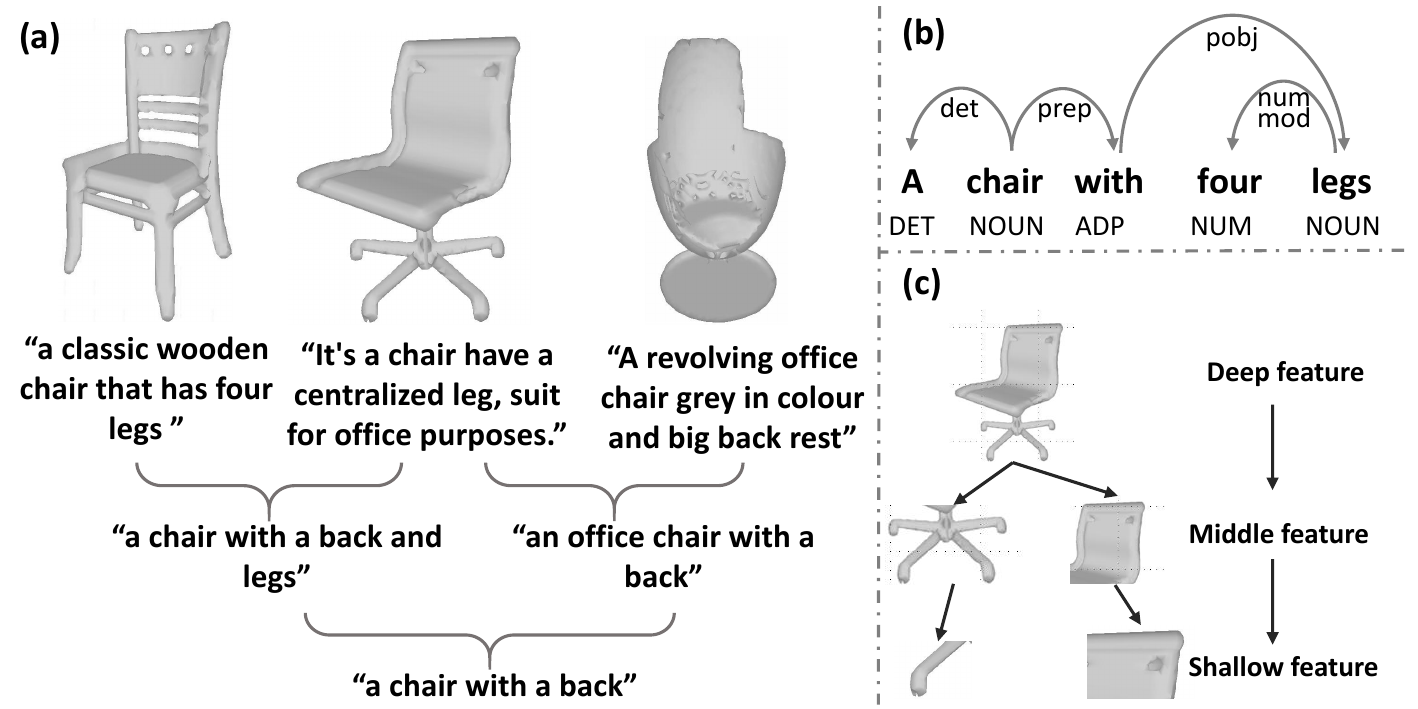}
   \caption{Hyperbolic text-shape representations. (a) The hierarchical structure between text and 3D shape. (b) The syntactic tree of text. (c) The hierarchical part-to-whole relationships of 3D shape.}
   \vspace{-4mm}
    \label{fig: cover_image}
\end{figure}

We argue that \emph{hierarchy}, preordering elements of a set in increasing complexity is fundamental to linking geometry and language as illustrated in~\cref{fig: cover_image}. 3D shapes inherently exhibit \emph{compositionality}~\cite{slim20233dcompat,naeem20223d}, possessing hierarchical part-to-whole relationships~\cite{montanaro2022rethinking}. 
On the other hand, language exhibits a hierarchical tree-like syntactic structure~\cite{chomsky2002syntactic,duvsek2016sequence,zhang2019syntax}, rooted in inter-word relationships. 
Recognizing these parallel hierarchical natures requires rethinking text-to-shape correspondence also within a similar hierarchical framework. For example, a general prompt like ``a chair" can correspond to thousands of 3D shapes. In contrast, a more detailed description like ``a wooden chair with armrests and four legs" narrows down the possible shapes to those with specific attributes.
Fully embracing and leveraging such hierarchical nuances can significantly improve the fidelity and specificity of generated shapes, making strides in the field of text-to-shape synthesis.

Existing text-to-shape methods can be divided into two categories depending on the data type they handle: one for paired text-shape data~\cite{text2shape2018,autosdf2022,shapecrafter2022,implicit2022,sdfusion2023,diffusionsdf2023,tian2023shapescaffolder} and the other for unpaired data~\cite{magic3d2023,dreamfusion2022,DreamBooth3D2023,podia2023,dream3d2023,score2023,sanghi2022clip,sanghi2023clip}. Some methods~\cite{magic3d2023,dreamfusion2022,DreamBooth3D2023,podia2023,dream3d2023,score2023} for unpaired data generate intermediary images, and then transform these images into 3D shapes by 3D generative models, like NeRFs~\cite{magic3d2023,dream3d2023}. Others~\cite{sanghi2022clip,sanghi2023clip} leverage a shared text-image embedding space, generating 3D shapes by an image-3D generator. These methods do not directly learn the text-shape representation. In contrast, methods using text-shape paired data have the advantage of directly learning the text-shape representation by GANs~\cite{text2shape2018}, Variational Autoencoders~\cite{shapecrafter2022,autosdf2022}, or Diffusion Models~\cite{sdfusion2023,diffusionsdf2023}, generating 3D shapes from texts. To date, all these methods ignore the joint hierarchical structure of 3D shapes and natural language. 

In this work, we focus on text-shape paired data. Inspired by prior works on leveraging hierarchies in images~\cite{imagehyperbolic2020,imagehyperbolic2022,ermolov2022hyperbolic}, point clouds~\cite{montanaro2022rethinking,lin2023hyperbolic}, or text-image pairs~\cite{imagetexthyperbolic2023}, we propose to embed the tree-like hierarchical structure of text and shape, jointly, into a more natural non-Euclidean, hyperbolic space. This incurs less distortion than in Euclidean space, primarily due to the exponential expansion of hyperbolic space ideally suited to representing trees~\cite{mettes2023hyperbolic}. Our approach, deemed \textbf{HyperSDFusion}, first utilizes a Signed Distance Field (SDF) based Autoencoder~\cite{sdfusion2023} to embed the SDF representation of 3D shapes into a compact latent space, learning a latent feature for each 3D shape. Then, to concurrently exploit the sequential (word order) and hierarchical structures of an input prompt, we propose a dual-branch latent diffusion model in hyperbolic space to generate a desired latent feature close to the ground truth latent feature from a noise. In one branch, we leverage a pre-trained text-image model~\cite{imagetexthyperbolic2023}, learning both sequential features of language and multi-modal hierarchical features of text-image, in hyperbolic space. In a parallel branch, we devise a hyperbolic text-graph convolution model that parses the input prompt into a syntax graph and learns the hierarchical features of language in hyperbolic space. Notably, to maintain the hierarchical structure of 3D shapes during the generation process, we introduce a hyperbolic hierarchical loss, which correlates the distance between 3D deep and shallow features with the distance to the origin of the Poincar\'{e} ball (hyperbolic space).

Finally, we conduct a series of experiments on the existing text-shape paired dataset, Text2Shape~\cite{text2shape2018}. The experiments demonstrate that our method achieves high-quality generation results while preserving the hierarchical characteristics of text and shape.
Our main contributions are:
\begin{itemize}
    \item We are the first to learn a joint hierarchical representation of text and shape in the hyperbolic space, improving the quality of text-to-shape generation.
    \item We introduce a dual-branch diffusion to fully capture both sequential and hierarchical structures of texts in hyperbolic space.
    \item Our proposed hyperbolic hierarchical loss ensures that the generation process of the diffusion model maintains the hierarchical structure of 3D shapes.
\end{itemize}

\begin{figure*}[htb]
    \centering
    \includegraphics[width=1.\linewidth]{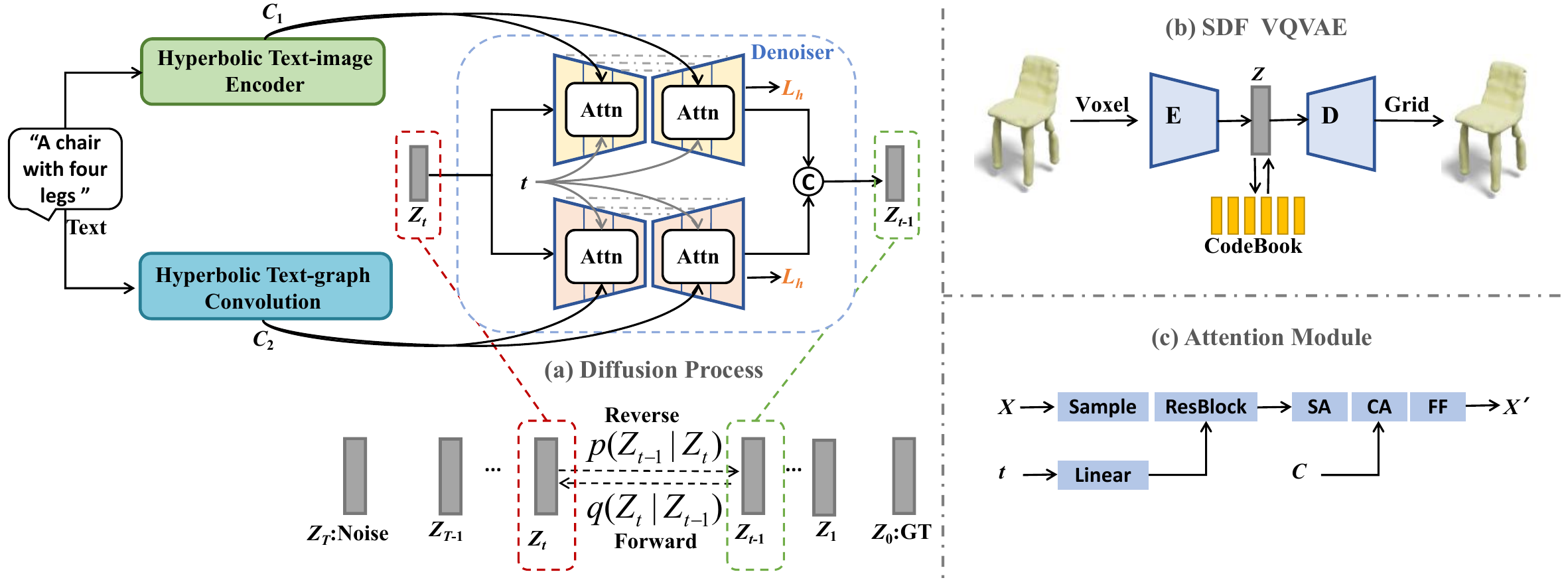}
    \caption{Overview of the proposed HyperSDFusion. \textbf{(a)} The forward and reverse processes of the proposed dual-branch diffusion model from $Z_{0}$ to $Z_{T}$. In particular, the detailed denoising process of the latent feature $Z_{t}$ based on text conditions $\{C_{1},C_{2}\}$ is showcased. \textbf{(b)} The architecture of a VQVAE for 3D shape represented by SDF. \textbf{(c)} The attention module in the denoiser of the diffusion model.\vspace{-3mm}}
    \label{fig: overview}
\end{figure*}

\section{Related works}
\paragraph{Text-to-Shape Generation.} In recent years, text-to-shape generation has garnered significant attention in 3D representation learning. In this task, annotating paired text-shape data is time-consuming and laborious. The Text2Shape dataset~\cite{text2shape2018}, a text-shape paired dataset, is widely used in text-to-shape generation. This dataset was proposed by Chen~\etal as the first text-shape paired dataset, and Chen~\etal implemented a GAN-based text-shape generation on it. Later, a series of methods~\cite{autosdf2022,shapecrafter2022,implicit2022,sdfusion2023,diffusionsdf2023,tian2023shapescaffolder} were proposed for paired data, which can be divided into VAE-based methods, Autoencoder-based methods, Auto-regressive-based method, and diffusion model-based methods according to the advanced models used. In VAE-based methods~\cite{shapecrafter2022,autosdf2022}, Fu~\etal~\cite{shapecrafter2022} propose ShapeCrafter, a recursive text-shape generation method by recursively embedding text features. In Autoencoder-based methods~\cite{implicit2022,tian2023shapescaffolder}, Tian~\etal~\cite{tian2023shapescaffolder} propose a structure-aware method to align the text feature space and the 3D shape feature space at the part level, by dividing the feature space. In Auto-regressive-based methods, Luo~\etal~\cite{luo2023learning} propose an improved Auto-regressive Model for 3D shape generation, by applying discrete representation learning in a latent vector instead of volumetric grids. In diffusion-based methods~\cite{sdfusion2023,diffusionsdf2023}, Cheng~\etal~\cite{sdfusion2023} propose SDFusion that employs latent diffusion to generate an ideal latent feature, close to the ground truth latent feature embedded by a Quantised-VAE.

Recently, with the development of multi-modal learning, researchers have used existing multi-modal models to generate 3D by unpaired text-shape data. Some methods~\cite{dream3d2023,magic3d2023,dreamfusion2022,DreamBooth3D2023} use images as intermediate generations, firstly leveraging pre-trained text-to-image models to produce images, then employ 3D reconstruction methods such as NeRFs to reconstruct 3D shapes from the images. Other methods~\cite{sanghi2022clip,sanghi2023clip} leverage a shared text-image embedding space, generating 3D shapes by an image-3D generator. However, these methods cannot directly learn the representation between text and 3D. In this work, we focus on 3D generation by paired text-shape data, aiming to directly learn the representation between text and 3D.

\paragraph{Diffusion Models.} Diffusion models as new powerful generative models have shown record-breaking performance in many applications, like 3D shape generation~\cite{hui2022neural,vahdat2022lion,zhang20233dshape2vecset,shim2023diffusion,zheng2023locally}, image synthesis~\cite{stablediffusion2022,dhariwal2021diffusion,wu2023harnessing}, human motion generation~\cite{zhang2022motiondiffuse,yuan2023physdiff,karunratanakul2023guided}, video generation~\cite{harvey2022flexible,ceylan2023pix2video}, etc. Diffusion models can be divided into two categories based on whether they directly generate the final output: standard diffusion models~\cite{zhang2022motiondiffuse,chen2023diffusiondet} and latent diffusion models~\cite{stablediffusion2022,peebles2023scalable}. The standard diffusion models directly generate the final output, such as images or 3D shapes. However, as the scale or resolution of the output increases, standard diffusion models consume significant GPU resources. The latent diffusion models utilize a learned latent space to generate a latent feature and then transform it into the final output, substantially reducing GPU consumption. In this work, we also employ the latent diffusion model to generate a latent feature, and then transform it into a 3D shape.

\paragraph{Hyperbolic Representation learning.} In recent years, there has been a growing interest in deep representation learning in hyperbolic spaces~\cite{peng2021hyperbolic,yang2022hyperbolic,zhou2023hyperbolic}. It has been proved that hyperbolic space is more suitable than Euclidean space for processing data with a tree-like structure or power-law distribution due to its exponential growth property. There are a few works about hyperbolic representation learning in Computer Vision~\cite{imagehyperbolic2020,imagehyperbolic2022,montanaro2022rethinking,ermolov2022hyperbolic,lin2023hyperbolic,imagetexthyperbolic2023,leng2023dynamic}, such as learning hierarchical representation of images in hyperbolic space~\cite{imagehyperbolic2020,imagehyperbolic2022,ermolov2022hyperbolic}, analyzing and utilizing the hierarchical property of point cloud in hyperbolic space~\cite{montanaro2022rethinking,lin2023hyperbolic}, etc. Desai~\etal~\cite{imagetexthyperbolic2023} presented the hierarchical structure in text-image and proposed a hyperbolic contrastive learning model for text-image paired data. As far as we know, we are the first to learn the hierarchical representation of text and shape in hyperbolic space, which is beneficial for text-to-shape generation.

\section{Preliminaries}
\label{sec:pre-thorey}
\paragraph{Hyperbolic space.} Hyperbolic space is a non-Euclidean space, also an n-dimensional Riemannian manifold of constant negative curvature. The hyperbolic space can be modeled by several isometric models~\cite{yang2022hyperbolic}. The most popular model is the \Poincare ball model, which we adopt. The $r$-dimensional \Poincare ball $B_{c}^{n}\vcentcolon=\{x\in\mathbb{R}^n \mid \|x\|^2<r^2\}$ with a negative curvature $c$, endowed with the canonical metric of the Euclidean space $g_{x}^{B}$ admits the structure of a Riemannian manifold $\mathcal{B}\vcentcolon=(B_{c}^{n},g_{x}^{B})$ with the geodesic distance $d(\cdot):\mathcal{B}\times\mathcal{B}\to\mathbb{R}$ given by:
\begin{align}
    d(x,y) = \frac{1}{\sqrt{c}}\mathrm{arcosh}\left(1+\frac{2\|x-y\|^2}{(1-\|x\|^2)(1-\|y\|^2)}\right).
\end{align}
Identifying its tangent space $\mathcal{T}_x\mathcal{B}$ with $\mathbb{R}^n$ allows us to transform data between Euclidean and hyperbolic spaces through the exponential and logarithmic maps with origin at $x$, respectively denoted by $\Exp_x$ and $\Log_x$. Explicit expressions are explained in Mettes~\etal ~\cite{mettes2023hyperbolic}.

\paragraph{Hyperbolic Graph Convolution (HGC)} HGC extends the standard graph convolution to hyperbolic spaces and is suitable for processing tree-like graph data. Generally, an HGC consists of feature initialization, feature updating and aggregation, and activation. Formally, a graph is defined as $\mathcal{G}=(\mathcal{V},\mathcal{E})$, where $\mathcal{V}=\{x_{i}^{E}\|i=0,...,N\}$ is the node feature set in Euclidean space, and $\mathcal{E}$ is the edge set. Firstly, an $\Exp$ function transfers node features to hyperbolic space, initializing node features on the hyperbolic manifold, $x_{i}^{\mathcal{B}}$. Then a $\Mobius$-layer~\cite{kochurov2020geoopt} updates and aggregates node features, which is a generalization of the fully connected layer in hyperbolic space. Finally, a non-linear hyperbolic activation $\sigma^{B}$ acts on these features by: (i) mapping them back to Euclidean space, (ii) transforming them by traditional (Euclidean) non-linear layers, and (iii) mapping them back to the hyperbolic space. This yields hyperbolic node features $y_{i}^{B}$. This procedure is defined as:
\begin{equation}
    y_{i}^{B}=\sigma ^{B}(\Mobius(\Exp(x_{i}^{B}))).
\end{equation}
We refer the reader to Yang~\etal~\cite{yang2022hyperbolic} for further details.

\begin{figure*}[htb]
    \centering
   \includegraphics[width=1\linewidth]{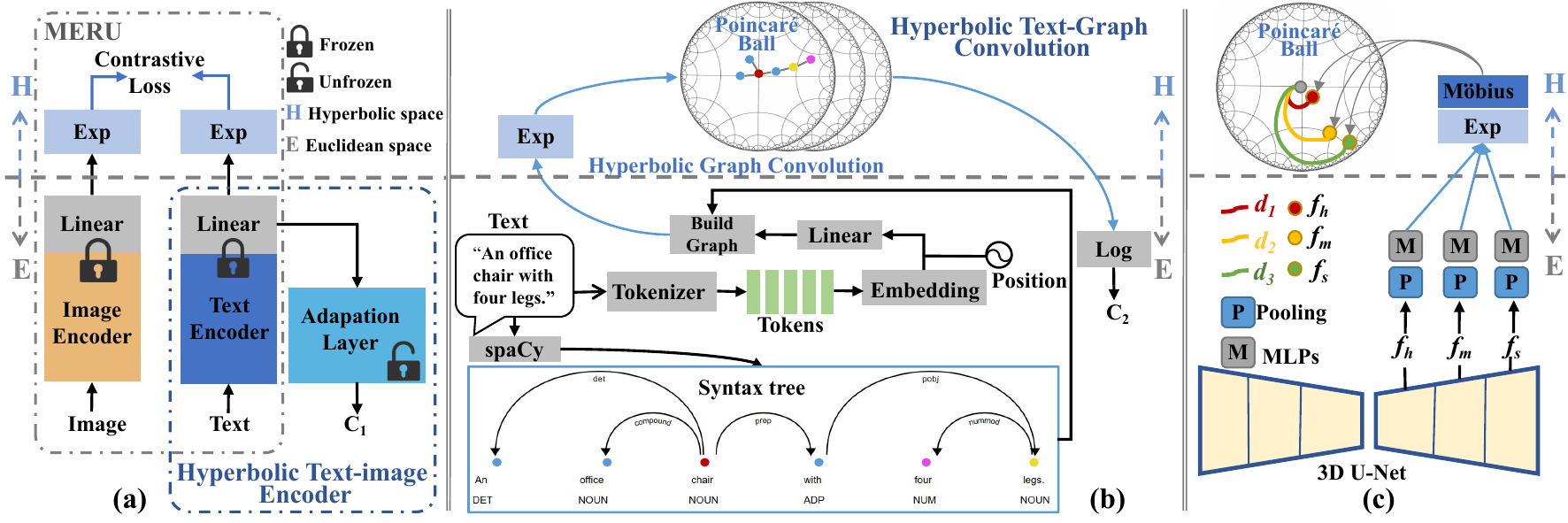}
   \caption{Illustration of our proposed modules. \textbf{(a)} Given a text, the hyperbolic text-image encoder learns both sequential and multi-modal hierarchical features of the text, $C_{1}$. \textbf{(b)} The hyperbolic text-graph convolution module learns hierarchical syntactic features of the text, $C_{2}$. \textbf{(c)} Hyperbolic hierarchical loss supervises the hierarchical structure of 3D shape features in hyperbolic space. \vspace{-4mm}
   }
    \label{fig: three}
\end{figure*}

\section{Method}
\label{sec:method}
In this section we present our method, called HyperSDFusion, for text-to-shape generation. Similar to the visual attribute hierarchies of Li~\etal~\cite{li2023euclidean}, we encode the semantic hierarchy in a hyperbolic space, where the root of the hierarchy (i.e., at the center of Poincar\'{e} ball) is a category, e.g., chair. The finer-grained separations are at lower levels, such as subcategories or details. Finally, the category-irrelevant features are at the lowest level, e.g., legs of a chair (\cf Fig.~\ref{fig: cover_image}).
We achieve this through hyperbolic text-shape feature learning under the supervision of our proposed hyperbolic hierarchical loss.
We now describe our model and provide additional details in our supplementary material.

\subsection{A Dual-branch Latent Diffusion Model}
The architecture of our proposed HyperSDFusion for text-to-shape generation is shown in~\cref{fig: overview}. HyperSDFusion is a dual-branch Latent diffusion model, including 3D shape compression, the forward and reverse process of the latent diffusion model based on text conditions.

\paragraph{3D Shape Compression} As the scale of the 3D shape increases, the GPU consumption yields a significant increment. Embedding the 3D shape into the low-dimensional latent space tends to greatly reduce resource consumption. Hence, we encode 3D shapes into a compact latent space, representing each 3D shape as a latent feature.

As shown in~\cref{fig: overview}{\color{red}(b)}, we firstly represent 3D shapes as a Truncated Signed Distance Field (TSDF) $\Gamma$ of size $R\times R\times R\times 1$, where $R$ is the resolution, and 1 is the dimension of distance. A Vector Quantised-Variational AutoEncoder (VQ-VAE)~\cite{van2017neural} is employed to learn the latent feature of $\Gamma$. The encoder $E$ of VQ-VAE encodes $\Gamma$ into a latent representation $Z=E(\Gamma)$, where $Z \in \mathbb{R}^{d\times d\times d\times m} $, $d$ is the resolution of the latent feature and $m$ is the dimension of features. Importantly, the encoder downsamples the TSDF by a factor $f=R/d$. Then, $Z$ is discretized by the codebook $VQ$. Finally, the decoder $D$ of VQ-VAE reconstructs the 3D shape from the discretized $Z$, represented as $\Gamma^{\prime }=D(VQ(Z))$. The reconstruction loss between $\Gamma$ and $\Gamma^{\prime }$ follows Van~\etal~\cite{van2017neural}.

\paragraph{Forward Process of The Latent Diffusion Model.} With the learned latent feature $Z$ of each shape as ground truth, the forward process is an iterative process that adds Gaussian noise to $Z$ in a Markovian manner, as shown in~\cref{fig: overview}{\color{red}(a)}. In detail, the ground truth latent feature of shape is the input at time $0$, defined as $Z_{0}$. Then $Z_{0}$ is noised by Gaussian noise $\epsilon$ time by time. After T times, $Z_{0}$ is noised to $Z_{T}$, which is close to standard Gaussian noise. The whole forward process is represented as:
\begin{equation}
    q(Z_{1:T}|Z_{0})=\prod_{t=1}^{T}q(Z_{t}|Z_{t-1}), 
\end{equation}
where $q$ is the feature distribution at each time.

\paragraph{Reverse Process of The Latent Diffusion Model based on Text Conditions.} Given a text prompt as the condition, the reverse process is to generate the latent feature of the 3D shape coincident with the text from random Gaussian noise. In detail, starting from random Gaussian noise $Z_{T}$ at time $T$, we gradually sample it to remove noise by a reverse Markov chain. As shown in~\cref{fig: overview}{\color{red}(a)}, the noisy latent feature at time $t$, $Z_{t}$, is denoised to $Z_{t-1}$ by a denoiser conditioned on text features, $C_{1}, C_{2}$. The denoiser is a dual-branch 3D U-Net structured attention model~\cite{stablediffusion2022}. Each module in the attention model consists of an up/down- sample, residual blocks, self-attention, cross-attention, and a feed-forward layer, as shown in~\cref{fig: overview}{\color{red}(c)}. After $T$ times, the output $Z_{0}$ is close to the ground truth latent feature. The reverse process is formulated as follows:
\begin{equation}
    p(Z_{0:t}|Z_{T})=p(Z_{T})\prod_{t=1}^{T}p(Z_{t-1}|Z_{t}),
\end{equation}
where $p$ is the feature distribution in the reverse process. 

\paragraph{Dual-branch Denoiser.} To well utilize text information, we design a dual-branch denoiser. On the one hand, text is composed of ordered words, indicating a sequential structure. We introduce a hyperbolic text-image encoder to learn the multi-modal sequential feature of the text, defined as $C_{1}$. On the other hand, natural language also exhibits a syntactic structure~\cite{chomsky2002syntactic} that can be parsed into a syntax tree, as shown in~\cref{fig: cover_image}{\color{red}(b)}. We propose a hyperbolic text-graph convolution module to learn the hierarchical structure of the text, defined as $C_{2}$. 

A common way to utilize the two conditions $C_{1}, C_{2}$ is to concatenate them as a single condition and inject them into a single-branch denoiser. However, this way may cause feature interference. Hence, we propose a dual-branch denoiser to utilize respectively, which consists of two parallel 3D U-Nets structures. As shown in~\cref{fig: overview}{\color{red}(a)}, $C_{1}$ and $C_{2}$ are fed separately as conditions to the cross-attention of the 3D U-Nets. Finally, the outputs of the two branches are concatenated. The dual-branch denoiser can preserve the independence of the two conditions, preventing confusion or loss of information. In this way, one branch perceives the sequential structure from $C_{1}$, while the other branch perceives the hierarchical structure from $C_{2}$.

\subsection{Text Feature Learning in Hyperbolic Space}
\label{sec: text learning}
As previously mentioned, text exhibits both sequential structure and syntactic hierarchical structure. We now expose our hyperbolic text-image encoder (HTIE) and hyperbolic text-graph convolution module (HTGC) used to better capture these structures.

\paragraph{Hyperbolic Text-image Encoder.} Our hyperbolic text-image encoder learns text features, not only capturing the sequential structure of text but also embedding the multi-modal hierarchical structure of text-image. As shown in~\cref{fig: three}{\color{red}(a)}, HTIE consists of a transformer-based text encoder and an adaptation layer. The text encoder captures long-range dependencies between words in a text, learning the sequence features of text. Besides, the text encoder is pre-trained from MERU~\cite{imagetexthyperbolic2023}, a text-image contrastive learning model that learns the hierarchical features between text and image in a hyperbolic space. Hence, our text encoder also embeds multi-modal hierarchical features of text and image. Then, we introduce an adaption layer to narrow the gap between the pre-trained text encoder and the text-shape dataset, implemented by multiple transformer layers. In this way, our HTIE learns a text condition, $C_{1}$.

\paragraph{Hyperbolic Text-graph Convolution Module.} A text can be parsed into a syntax tree according to its syntactic structure. As shown in~\cref{fig: three}{\color{red}(b)}, each word is endowed with Part-Of-Speech (POS), and the dependency between words is indicated by directed edges. Since hyperbolic space is more suitable for processing tree-like structure data than Euclidean space~\cite{yang2022hyperbolic}, we propose a hyperbolic text-graph convolution module to learn the syntactic structure of text, including syntax tree construction, text-graph initialization in Euclidean space, and learning in hyperbolic space. 

Firstly, the input text is processed by spaCy~\cite{spacy2}, a natural language processing library yielding a syntax tree. Secondly, the node and edge features are initialized in the Euclidean space $E$. A given text prompt is converted into tokens and then mapped to a vector representation by an embedding layer. The vector representation is added with the position encoding, following a linear layer to transform features. At this point, the features of tokens in Euclidean space are initialized as node features $V^{E}$ of the text-graph, and the branches of the syntax tree serve as edges $\mathcal{E}$ in the text-graph. The text-graph is defined as $\mathcal{G}^{E}=\{\mathcal{V}^{E},\mathcal{E}\} $. Thirdly, the text-graph is projected to hyperbolic space, learning the hierarchical structure of the text-graph. In detail, the projected text-graph is represented as $\mathcal{G}^{H}=\Exp(\mathcal{G}^{E})=\{\mathcal{V}^{H},\mathcal{E}\}$, where $\Exp$ is as defined in Section~\ref{sec:pre-thorey}, and $H$ represents hyperbolic space. In hyperbolic space modeled by the \Poincare Ball model, stacked hyperbolic graph convolutions propagate and update node features on the text-graph, better capturing the hierarchical structure of the text-graph. At last, the updated node features of the text-graph are projected back into the Euclidean space by the $\Log$, yielding the other text condition $C_{2}$.

\subsection{Hyperbolic Hierarchical Loss for 3D shape}
\label{sec:loss}
In this work, the denoiser predicts the latent feature of 3D shapes from a random Gaussian noise conditioned on text, denoted as $Z^{\prime}$. To ensure that $Z^{\prime}$ closely approximates the ground-truth latent feature, the 
mean squared error $L_{mse}$ between $Z$ and $Z^{\prime}$ is employed as the loss function. However, $L_{mse}$ only supervises the similarity between features, ignoring the hierarchical structure of 3D shapes.

As mentioned before, 3D shapes inherently exhibit a hierarchical structure, namely the part-whole hierarchy. In the feature space of 3D shape, this hierarchy manifests as hierarchical relationships between deep and shallow features. In this work, these deep and shallow features are the multi-scale outputs of the 3D U-Net. As depicted in~\cref{fig: three}{\color{red}(c)}, the small-scale features $f_{h}$ are the global feature of the 3D shape, the large-scale features $f_{s}$ are the local feature, and the medium-scale features $f_{m}$ line in between. To supervise these features to maintain the hierarchy, we propose a hyperbolic hierarchical loss to regularize the feature space.

These features are first passed through a pooling layer and multi-layer perceptrons to unify their scale and feature dimension. Then, they are projected to the hyperbolic space by $\Exp$, and transformed to a unified distribution by a shared $\Mobius$ layer. Due to the hierarchical property of deep and shallow features, features in the hyperbolic space should also maintain a tree-like hierarchical structure, as illustrated in~\cref{fig: three}{\color{red}(c)}. Here, we use the relative distance between features to constrain the feature distribution, i.e., $d_{2}>d_{1}$, $d_{3}>d_{2}$. $d_{i}$ is the geodesic distance from the $i$-th feature to the center point. Therefore, the hyperbolic hierarchical loss is defined as:
\begin{equation}
L_{h}=\max\left ( 0,- d_{2}+d_{1} \right ) + \max\left ( 0, - d_{3}+d_{2} \right ).
\end{equation}

The whole training loss is given by $L=L_{mse}+\alpha  L_{h}$, where $\alpha$ is a balancing factor.

\section{Experiments}
To evaluate the performance of our method, we conducted a series of experiments on the paired text-shape dataset, Text2Shape~\cite{text2shape2018}. The details of the experiments are described in the following subsections.

\paragraph{Implementation Details.}
Our method generates a 3D shape from a text, which is represented as a TSDF with the size of $64\times 64 \times 64\times 1$, where 64 is the resolution. Our method includes a 3D VQ-VAE and a dual-branch diffusion model. Firstly, we trained the 3D VQ-VAE on the ShapeNet dataset~\cite{shapenet} with 13 categories of 3D shapes. The 3D VQ-VAE compresses the TSDF to a compact latent feature $Z$ with the size of $16\times 16\times 16\times 3$. Then, the 3D VQ-VAE is frozen, and the dual-branch diffusion model is trained on the Text2Shape dataset~\cite{text2shape2018}. The optimizer for the training is AdamW with an initial learning rate of 1e-5. During inference of the diffusion model, the sampler is the DDIM sampler~\cite{ddim}, where the sample step $T$ is set to 100. 

\paragraph{Dataset.} In this work, we choose Text2Shape dataset~\cite{text2shape2018} to conduct experiments. The reason is that this work aims to learn hierarchical representations of text and 3D directly, which requires paired data. Text2Shape dataset~\cite{text2shape2018} is a widely used dataset in paired text-to-shape generation. This dataset provides rich natural language annotations for tables and chairs in ShapeNet~\cite{shapenet}, describing their shape, color, texture, material, etc. This work only focuses on how to generate shapes from text.

\paragraph{Evaluation Metrics.}
In order to assess the generation quality, we adopt Intersection over Union (IoU), Chamfer Distance (CD), F-score, and Fr\'{e}chet Inception Distance (FID) as evaluation metrics. IoU computes the intersection volume between generated and GT 3D shapes. CD is computed between generated and GT point clouds that are sampled from 3D shapes by farthest point sampling. F-score computes the harmonic mean between precision and recall, based on the distance between generated 3D shapes and GT 3D shapes. We set the distance threshold is set to $1\%$, as in Tatarchenko~\etal~\cite{tatarchenko2019single}. FID calculates the Fr\'{e}chet distance between the feature representations of images rendered from predicted and GT 3D shapes.

For evaluating the hierarchy of generated 3D shapes, we introduce two metrics, Hierarchical Mutual Difference (HMD) and Hyperbolic Distance (HD). HMD computes the Chamfer distance between 3D shapes generated from a general text and a detailed text, assessing the text-shape hierarchical structure. HD computes the geodesic distance in hyperbolic space among the deep and shallow features of 100 randomly selected generated 3D shapes, evaluating the hierarchical structure of 3D shapes.

\begin{figure}[tb]
    \centering
    \includegraphics[width=0.9\linewidth]{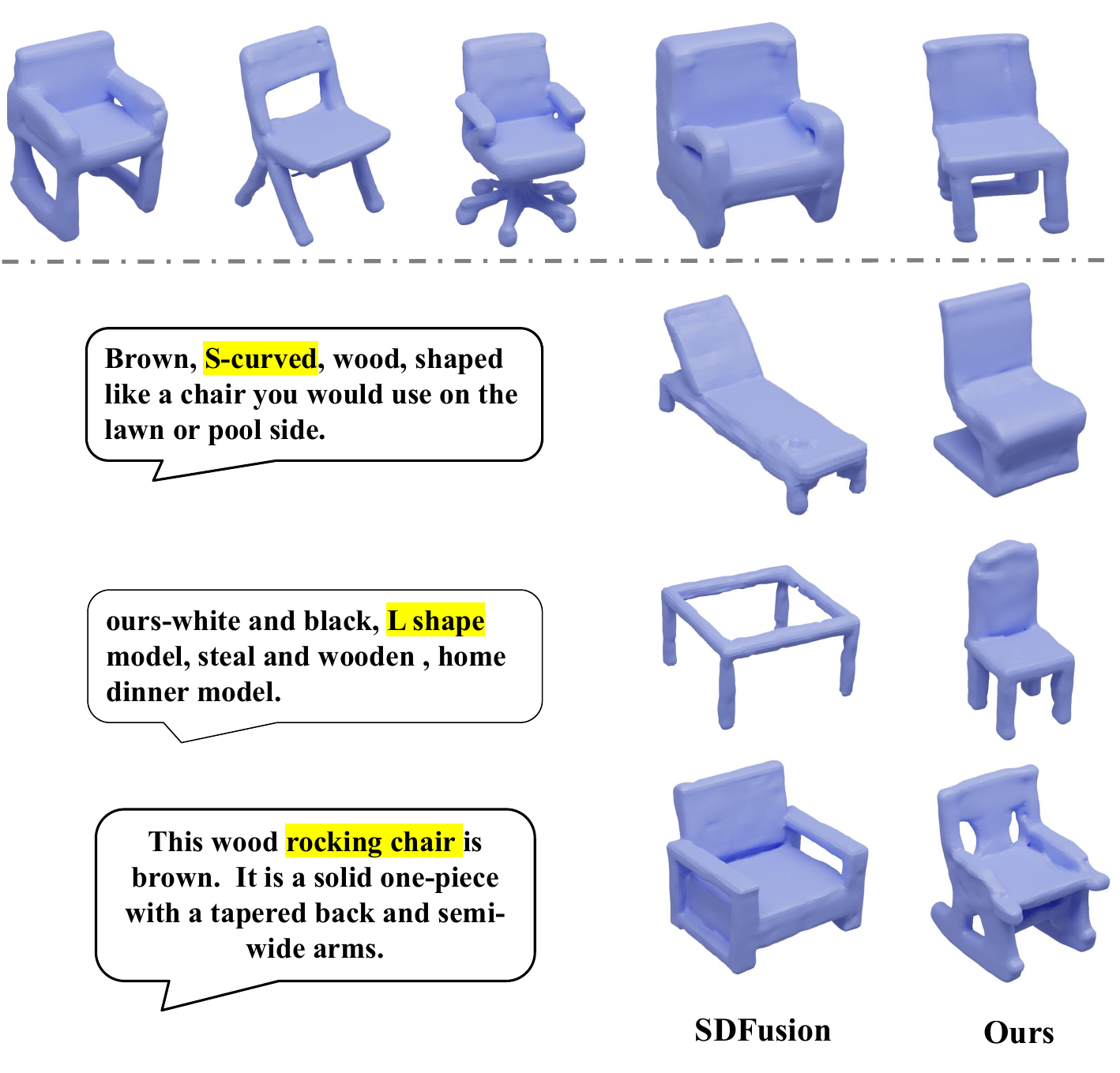}
    \caption{The showcase of text-to-shape generation results. Above the dotted line are some examples generated by our method, and below is the result compared to SDFusion~\cite{sdfusion2023}.\vspace{-2mm}}
    \label{fig:vis_sota}
\end{figure}

\begin{table}[tp]
\centering
\setlength{\tabcolsep}{10.5pt}\caption{Evaluating text-to-shape generation on random 1000 samples of Text2Shape~\cite{text2shape2018} test set. Bold indicates the best results.\vspace{-1mm}}
\resizebox{\linewidth}{!}{
\begin{tabular}{lccll}
\hline
Methods    & IoU$\uparrow$   & CD$\downarrow$   & F-score$\uparrow$ & FID$\downarrow$   \\ \hline
Liu~\etal~\cite{implicit2022}& 12.21  & 1.41   & 13.34            & 1.00         \\
SDFusion~\cite{sdfusion2023} & 13.98     & 1.246        & 12.67       & 5.47        \\
HyperSDFusion(Ours)   & \textbf{16.21}    & \textbf{0.7433}    & \textbf{15.16}   & \textbf{0.70}               \\ \hline
\end{tabular}\vspace{-2mm}}
\label{fig: compare with others}
\end{table}

\subsection{Text-to-Shape Generation Results}

\paragraph{Method for Comparison.} There are few existing diffusion-based methods for paired text-to-shape generation, including SDFusion~\cite{sdfusion2023} and Diffusion-SDF~\cite{diffusionsdf2023}. SDFusion learns the mapping between a 3D latent space and text feature space, while Diffusion-SDF is for patch-wise latent spaces. Because our method aims to learn the hierarchical representation in a joint text-shape latent space, we compare our method with SDFusion~\cite{sdfusion2023} in the experiment. SDFusion learns text features in Euclidean space, and generates the 3D shape without hierarchical supervision.

\paragraph{Results on Text-to-shape Generation.}~\cref{fig: compare with others} compares our HyperSDFusion against the previous SOTA text-to-shape generation method, SDFusion~\cite{sdfusion2023}. Quantitatively, our method outperforms SDFusion by a significant margin ($87\%$ decrease in FID, $40\%$ decrease in CD, $23\%$ improvement in IoU, and $19.7\%$ gain in F-score.). We achieved the best results on all these metrics for assessing generation quality. In particular, the large drop in FID indicates that shapes generated by our method are visually closer to the desired ones indicating high-quality text-to-shape generation. 
Some samples generated by our method are shown in~\cref{fig:vis_sota}. We observe that these shapes are complete and crisp in detail, when compared to SDFusion, which performs poorly on long texts. In contrast, our approach adequately captures text features and generates shapes faithful to the text, such as S-shapes, L-shapes, and rocking chairs. 

\paragraph{Generation Performance on Texts with Different Lengths.} For evaluating the generation performance on texts with different lengths, we conducted a comparison on Text2Shape++ dataset~\cite{shapecrafter2022}, which is built on Text2Shape dataset. In Text2Shape++, each text prompt is represented as a phrase sequence, and each phrase sequence corresponds to one or more shapes. We generate shapes with texts of length less than 8, more than 8 less than 16, and more than 16, respectively, to compare the performance of our method with SDFusion~\cite{sdfusion2023}. The comparison results are listed in~\cref{tab: compare on length}. It can be observed that our method outperforms the existing methods on both long and short texts. We also showcase more generations of texts with different lengths in~\cref{fig:more_vis}{\color{red}(a)}. Our method generates shapes consistent with both short and long texts. These results indicate that our method sufficiently learns and utilizes the text feature.

\begin{table}[tp]
\centering
\caption{Comparsion results on texts with different lengths.\vspace{-2mm}}
\resizebox{\linewidth}{!}{
\begin{tabular}{l|cccc|cccc|cccc}
\hline
\multirow{2}{*}{Method} & \multicolumn{4}{c|}{$words \leq 8$}                                                             & \multicolumn{4}{c|}{$8<words\leq 16$}                                                          & \multicolumn{4}{c}{$16<words$}                                                            \\ \cline{2-13} 
                        & \multicolumn{1}{c|}{IoU$\uparrow$} & \multicolumn{1}{c|}{F-sc.$\uparrow$} & \multicolumn{1}{c|}{CD$\downarrow$} & FID$\downarrow$  & \multicolumn{1}{c|}{IoU$\uparrow$} & \multicolumn{1}{c|}{F-sc.$\uparrow$} & \multicolumn{1}{c|}{CD$\downarrow$} & FID$\downarrow$  & \multicolumn{1}{c|}{IoU$\uparrow$} & \multicolumn{1}{c|}{F-sc.$\uparrow$} & \multicolumn{1}{c|}{CD$\downarrow$} & FID$\downarrow$  \\ \hline
SDFusion~\cite{sdfusion2023}                & 8.52                     & 12.15                        & 1.71                    & 7.51          & 8.04                     & 10.28                        & 1.92                    & 8.19            & 10.13                    & 12.01                        & 1.48                    & 6.59          \\
Ours                    & \textbf{15.41}           & \textbf{13.58}               & \textbf{0.99}           & \textbf{3.64} & \textbf{14.56}          & \textbf{15.37}          & \textbf{0.81}           & \textbf{4.22}           & \textbf{13.25}           & \textbf{16.11}               & \textbf{0.71}           & \textbf{4.91} \\ \hline
\end{tabular}
\vspace{-2mm}}
\label{tab: compare on length}
\end{table}

\begin{table}[tp]
\centering
\caption{Ablations on architectures and loss functions. \textcircled{\footnotesize{1}} refers to the single-branch diffusion model. \textcircled{\footnotesize{2}} represents the dual-branch diffusion model. In bold indicates the best results.\vspace{-1mm}}
\resizebox{\linewidth}{!}{
\begin{tabular}{clccc}
\hline
Group              & Model                                    & IoU$\uparrow$   & CD$\downarrow$    & F-score$\uparrow$ \\  \hline
\multirow{1}{*}{0} & Baseline                    & 12.74        & 11.04            & 8.93                \\ \hline
\multirow{3}{*}{1}&Baseline+T5                    & 12.26        & 11.29            & 8.14                \\
                           &Baseline+CLIP                  & 13.99        & \textbf{9.41}            & 8.77                \\
                           &Baseline+HTIE                 & \textbf{14.26} & 10.49 & \textbf{9.42}    \\  \hline
\multirow{3}{*}{2}&Baseline+HTIE+HTGC+\textcircled{\footnotesize{1}} & 14.18         & 10.45           & 8.988               \\
                           &Baseline+HTIE+HTGC+\textcircled{\footnotesize{2}} &  \textbf{15.73} & \textbf{8.99}  & \textbf{9.50}  \\
                           &Baseline+HTIE+GCN+\textcircled{\footnotesize{2}}  & 15.41 & 9.28  & 9.19    \\ \hline
\multirow{2}{*}{3}&Baseline+HTIE+HTGC+\textcircled{\footnotesize{2}}+$L_{e}$  & 13.51 & 11.72 & 9.79         \\ 
                            &Baseline+HTIE+HTGC+\textcircled{\footnotesize{2}}+$L_{h}$   & \textbf{16.44}   & \textbf{9.053}           & \textbf{9.917}       \\ \hline
\end{tabular}
\vspace{-4mm}}
\label{tab: aba}
\end{table}
\vspace{-4mm}
\subsection{Ablation Studies}
We conducted ablation studies on a mini-set of Text2Shape dataset to demonstrate the effectiveness of our proposed method. We choose SDFusion~\cite{sdfusion2023} as the baseline because the text encoder of SDFusion is the standard sequential model in Euclidean space, Bert~\cite{bert}, and its denoiser is without the hierarchical supervision.

\begin{figure}
    \centering
    \includegraphics[width=\linewidth]{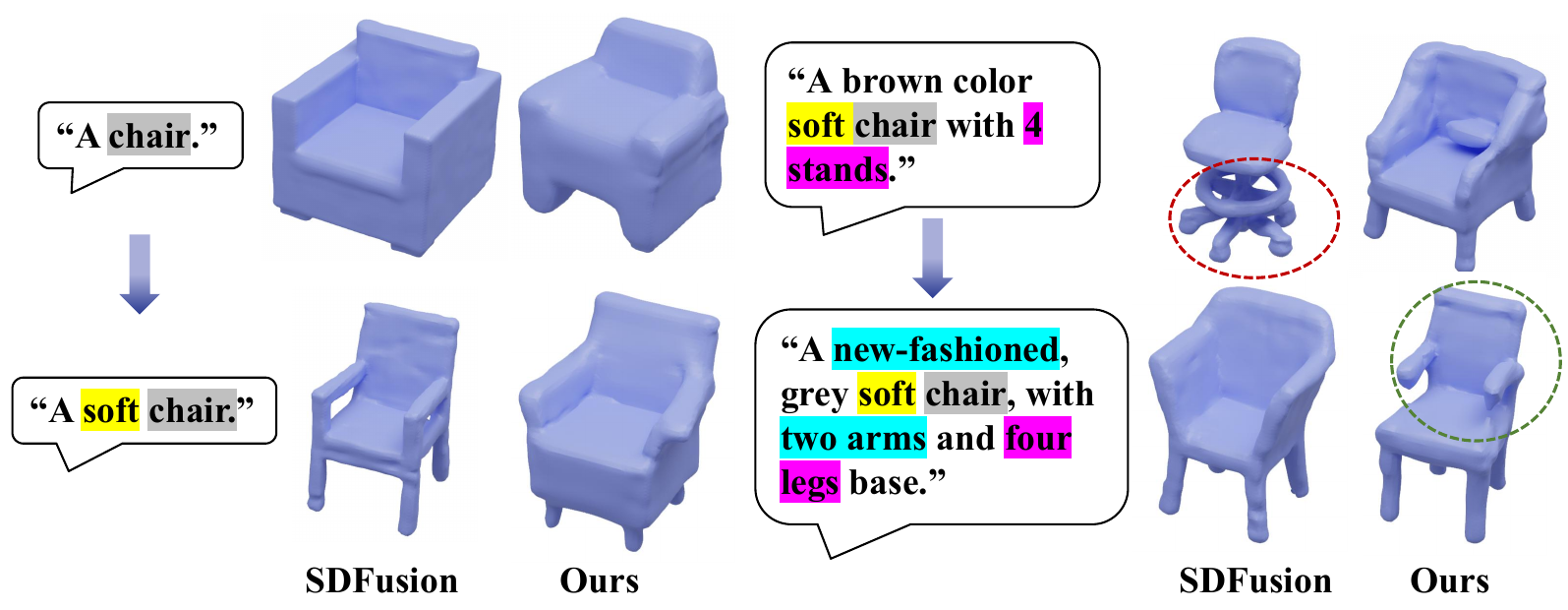}
    \caption{Visualizing text-shape hierarchical structure. Highlighted parts of the prompt represent the detailed information.\vspace{-2mm}}
    \label{fig:text-shape-hei}
\end{figure}

\begin{table}[tb]
\caption{The result of comparing the performance of capturing the text-shape hierarchical structure. More results are in the supplemental material.\vspace{-2mm}}
\resizebox{\linewidth}{!}{
\begin{tabular}{lll}
\hline
\multicolumn{1}{c}{\multirow{2}{*}{Text}} & \multicolumn{2}{c}{HMD$\downarrow$ }                                \\
\multicolumn{1}{c}{}                      & \multicolumn{1}{c}{SDFusion~\cite{sdfusion2023}} & \multicolumn{1}{c}{Ours} \\
\hline
``A chair."                                   &            -                  &              -           \\
``A soft chair."                    &       1.29            &     \textbf{0.57}          \\
``A brown color soft chair with 4 stands."    &        0.70      &      \textbf{0.20}          \\
``A new-fashioned, grey soft chair, with two arms and four legs base."     &      0.60    &     \textbf{0.43}      \\
\hline
\end{tabular}
\vspace{-4mm}}
\label{tab: text-shape hei}
\end{table}

\paragraph{Hyperbolic Text-image Encoder.} We compare our HTIE with the text encoder of the baseline, which is most used in existing methods~\cite{autosdf2022,sdfusion2023,diffusionsdf2023}. We replace the text encoder of the baseline with our HTIE. As listed in Group 1 of~\cref{tab: aba}, the baseline with our HTIE significantly improves the quantitative results, both IoU, CD, and F-score, indicating that our HTIE enhances the generation quality, owing to the multi-modal hierarchical features learned by the pre-trained MERU model.

In order to further verify the performance of our HTIE, we compare it with the other common text encoder, T5~\cite{T5}, and the Euclidean text-image encoder, CLIP~\cite{Clip}. Results in Group 1 of Tab.~\ref{tab: aba} show that T5 performs worse than CLIP and our HTIE. This is because T5, pre-trained on text-to-text generation, learns embedding in favor of text generation over contextual understanding. Moreover, our HTIE performs better than CLIP, manifesting that hyperbolic space is more suitable for text-image multi-modal learning.

\paragraph{Dual-branch Diffusion Model.} Our HTIE and hyperbolic text-graph convolution module (HTGC) learn different text features. We compare two ways of utilizing these two text features in diffusion models, single-branch and dual-branch. The comparison results are shown in Group 2 of~\cref{tab: aba}. The performance of the single-branch slightly decreases on IoU and F-score. The reason is that concatenating features of the single-branch lead to feature interference. In contrast, our dual-branch diffusion model further improves the performance of the model with HTIE, illustrating that our dual-branch architecture more effectively leverages the text features captured by HTIE and HTGC. Learned text features are visualized in the supplementary.

In addition, we compare our HTGC with the standard Euclidean graph convolution (GCN), as listed in Group 2 of~\cref{tab: aba}. It can be observed that our method yields better results, demonstrating hyperbolic space is more suitable for learning the syntactic structure of texts. 

\paragraph{Hyperbolic Hierarchical Loss.} We compared the performance of our model with and without hyperbolic hierarchical loss. As shown in Group 3 of~\cref{tab: aba}, the model with the hyperbolic hierarchical loss yields improvements in IoU and F-score while remaining comparable in CD, compared without the loss. It suggests that supervising the hierarchical structure of features during the denoising process contributes to improving the generation quality.

As listed in Group 3 of~\cref{tab: aba}, we also compare our hyperbolic hierarchical loss with an Euclidean version computed by Euclidean distance, $L_{e}$. The results show that $L_{h}$ is better, which reflects the advantage of hyperbolic space in maintaining the hierarchical structure.

\subsection{Analysis of Hierarchical Learning}
\label{sec: vis hierarchy}
\begin{figure}
    \centering
    \includegraphics[width=1.\linewidth]{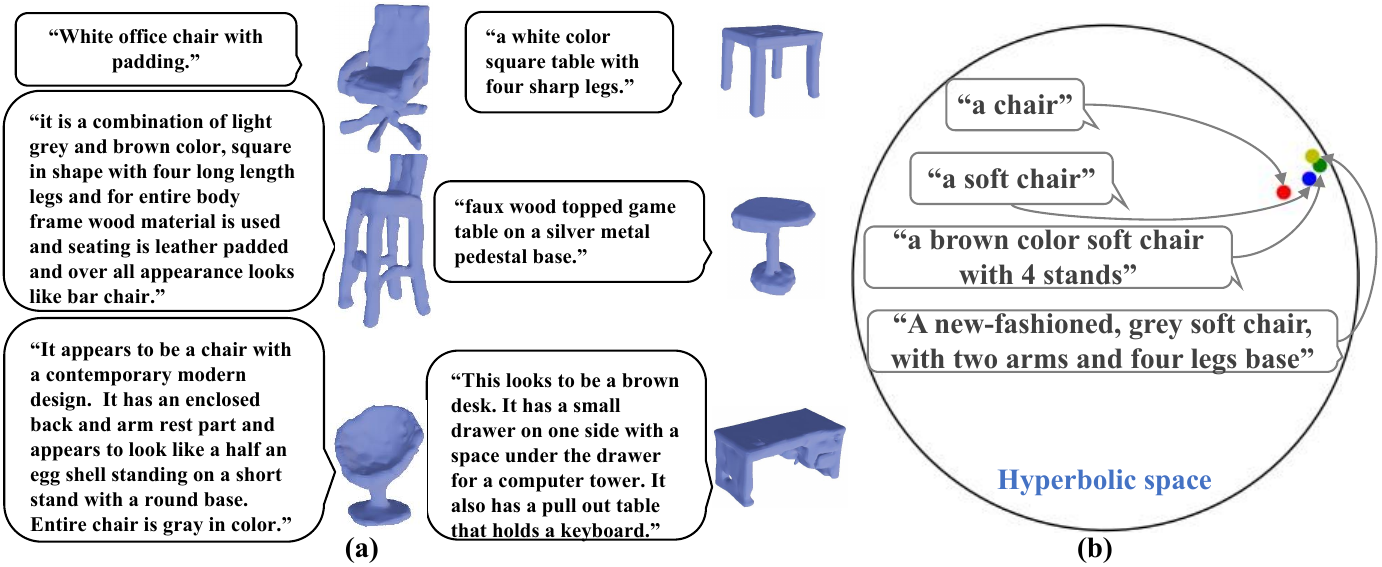}
    \caption{\textbf{(a)} More generation results, especially generated from long and complex text. \textbf{(b)} 3D Shape features generated from general and detailed texts in \Poincare Ball.\vspace{-4mm}}
    \label{fig:more_vis}
\end{figure}

\vspace{-1mm}
\paragraph{Analysis for Text-shape Hierarchical Structure.} Text-shape exhibits hierarchical structure from general texts and detailed texts. Our method embeds the hierarchical structure by hyperbolic learning, like our HTIE and HTGC. We use HMD to qualitatively evaluate the ability to capture the text-shape hierarchical structure. As shown in~\cref{tab: text-shape hei}, we enumerate the HMD of SDFusion~\cite{sdfusion2023} and our method on texts, ranging from general text to detailed text. We can observe that our method achieves smaller HMD between 3D shapes generated from different levels. It indicated that the shape generated from general text has hierarchical relationships with the shape generated from detailed texts. The hierarchical relationships are showcased in~\cref{fig:text-shape-hei}. Compared with SDFusion, given a general text, ``a chair", our method learned a shape without much detail. Indeed, a general text has no specific information, just like the root of the text-shape tree. Then adding the word ``soft" to the text, the chair generated by our method looks softer. Finally, by adding more detailed words, like ``four legs", ``new-fashion", ``two arms", our method generated accurate shapes with natural wrappings. We also visualized 3D shape features generated from general and detailed texts in hyperbolic space. Dots in~\cref{fig:more_vis}{\color{red}(b)} represent 3D latent shape features generated by texts. The hierarchical distribution of these dots is consistent with the text hierarchy.

\begin{table}[tb]
\centering
\setlength{\tabcolsep}{10.5pt}
\resizebox{\linewidth}{!}{
\begin{tabular}{lcccc}
\hline
Method   & $d_{1}$  &$d_{2}$  &$d_{3}$  &Order\\  \hline
SDFusion\cite{sdfusion2023} & 406.29   &431.76   &255.83   &$d_{2}>d_{1}>d_{3}$  \\
Ours     & \textbf{3.04}     &\textbf{3.44}     &\textbf{6.06}     &$d_{3}>d_{2}>d_{1}$  \\ \hline
\end{tabular}}
\caption{Comparing the performance in maintaining the hierarchical structure of 3D shape.\vspace{-3mm}}
\label{tab: 3d shape hei}
\end{table}

\paragraph{Analysis of the Hierarchical Structure of 3D Shape.} 
We employ a hyperbolic hierarchical loss to regularize the feature space in the denoiser. We use HD to qualitatively evaluate the ability to maintain the hierarchical structure of 3D shapes. As shown in~\cref{tab: 3d shape hei}, $d_{1}$, $d_{2}$, and $d_{3}$ are the distances of deep, middle, and shallow features to the origin in Euclidean space and hyperbolic space. If we compute the HD, the order provided by SDFusion is $d_{2}>d_{1}>d_{3}$, which does not follow the hierarchical structure of the point cloud. Instead, the order computed by our method is $d_{3}>d_{2}>d_{1}$, correctly following the hierarchy from deep to shallow features. It indicates the denoiser supervised by our hyperbolic hierarchical loss guarantees the tree-like hierarchical structure of 3D shape. We also provide the visualization of the feature distribution in the supplementary materials.

\section{Conclusion}
We propose a hyperbolic learning method for text-to-shape generation, namely HyperSDFusion. The key innovation lies in learning the inherent hierarchical structure of text and shape in hyperbolic space. In detail, we introduce a dual-branch diffusion model to fully utilize sequential and hierarchical features of text. The sequential features of the text are captured by the designed hyperbolic text-image encoder, simultaneously embedding multi-modal image-shape features. A hyperbolic text-graph convolution module is devised for learning hierarchical text features. Additionally, we propose a hyperbolic hierarchical loss to impart generated 3D shapes with hierarchical structure. Experimental results of our method on the Text2Shape dataset demonstrate the advantage of our method on text-to-shape generation. In the future, we will investigate more direct links to bridge language and 3D geometry.

\vspace{1mm}
\paragraph{Acknowledgments}
This work was supported by the National Nature Science Foundation of China under Grant 62272019, in part by China Scholarship Council. T. Birdal acknowledges support from the Engineering and Physical Sciences Research Council [grant EP/X011364/1].

\newpage
{
    \small
    \bibliographystyle{ieeenat_fullname}
    \bibliography{main}
}

\clearpage
\setcounter{page}{1}
\maketitlesupplementary

Our main paper introduced HyperSDFusion for text-to-shape generation, which explores how to bridge hierarchical structures in language and geometry. In this supplemental document, we provide more detailed information about our method and experiments.

\subsection{The Details of MERU}
In our hyperbolic text-image encoder introduced in Section~\ref{sec: text learning}, we employ the text encoder of MERU~\cite{imagetexthyperbolic2023} to learn text sequential features embedded with hierarchical multi-modal features. In this supplemental document, the details of MERU are described.

MERU is a large-scale contrastive image-text model that yields hyperbolic representations capturing the visual-semantic hierarchy. As shown in Figure~\ref{fig: three}, MERU consists of two separate text-image encoders, feature projection, and contrastive loss. The text encoder is multiple layers of transformer encoder blocks. The image encoder is the small Vision Transformer~\cite{chenempirical}. The feature projection is implemented by the $Exp$ function, which projects features to hyperbolic space. Under the supervision of the designed contrastive loss (a contrastive loss and an entailment loss), MERU enforces partial order relationships between paired text and images. For more details, please refer to ~\cite{imagetexthyperbolic2023}.

\subsection{The Details of Text Graph Building} The first step of our hyperbolic text-graph convolution module is text-graph initialization. In this supplemental document, we will explain more details of text-graph building.

As mentioned in Section~\ref{sec: text learning}, we process texts using spaCy~\cite{spacy2}, and obtain a syntax tree. The syntax tree is represented as a text graph by traversing the child nodes of the tree. The algorithm for the traversal process is elaborated in Alg~\ref{alg:tree-to-graph}.

\begin{algorithm}
    \caption{Framework of The Transformation of Tree-to-graph.}
    \label{alg:tree-to-graph}
    \KwIn{The syntax tree with $n$ nodes: $T_{G}=\{t_{i,G}|i=0,..,n-1\}$.}
    \KwOut{The adjacent matrix of a text graph: $M_{n\times n}$}
    \BlankLine
    $i = 0$;
    
    \While{$t_{i,G}$ in $T_{G}$}{
    $M_{i,i}=1$;
    
        \ForEach{child in $t_{i,G}$.child}{
            $j= child.index$;
            
          \If{ j $<$ n-1}{
            $M_{i,j}=1$;
            
            $M_{j, i}=1$;}
        }
    }    
\end{algorithm}

\subsection{The Details of Hyperbolic Hierarchical Loss}
As mentioned in Section~\ref{sec:loss}, we proposed a hyperbolic hierarchical loss to supervise the hierarchical structure of 3D feature space between deep and shallow features, $f_{h},f_{m}, f_{s}$, which are the output of the 3D U-Net at three scales. We process these features by our hyperbolic hierarchical loss, followed by steps shown in Alg~\ref{alg:loss}.

\begin{algorithm}
    \caption{Framework of Hyperbolic Hierarchical Loss.}
    \label{alg:loss}
    \KwIn{Deep features: $f_{h}$; \\ Middle features: $f_{m}$; \\ Shallow features: $f{s}$;\\ The dimensions of hyperbolic space: C.}
    \KwOut{Computed loss: $L$.}
    \BlankLine
    $f_{h}$ = MLP(Pooling($f_{h}$)); \\
    $f_{m}$ = MLP(Pooling($f_{m}$)); \\
    $f_{s}$ = MLP(Pooling($f_{s}$)); \\
    \ForEach{f in \{$f_{h},f_{m},f_{s}$\}}{
        f = Exp($\Mobius$(f));
    }
    
    ball = PoincareBall(c=1.0, dim=C); \\
    $d_{1}$ = ball.dist0($f_{h}$); \\
    $d_{2}$ = ball.dist0($f_{m}$); \\
    $d_{3}$ = ball.dist0($f_{s}$); \\
    L = max(0, -$d_{2}$+$d_{1}$) + max(0, -$d_{3}+d_{2}$).
    
\end{algorithm}

\subsection{More Qualitative Results on Capturing Text-shape Hierarchy}
In Section~\ref{sec: vis hierarchy}, we have given some results to present our advantage of capturing the text-shape hierarchy. In this supplemental document, we provide more qualitative results on capturing text-shape hierarchy.

\begin{figure}
    \centering
    \includegraphics[width=\linewidth]{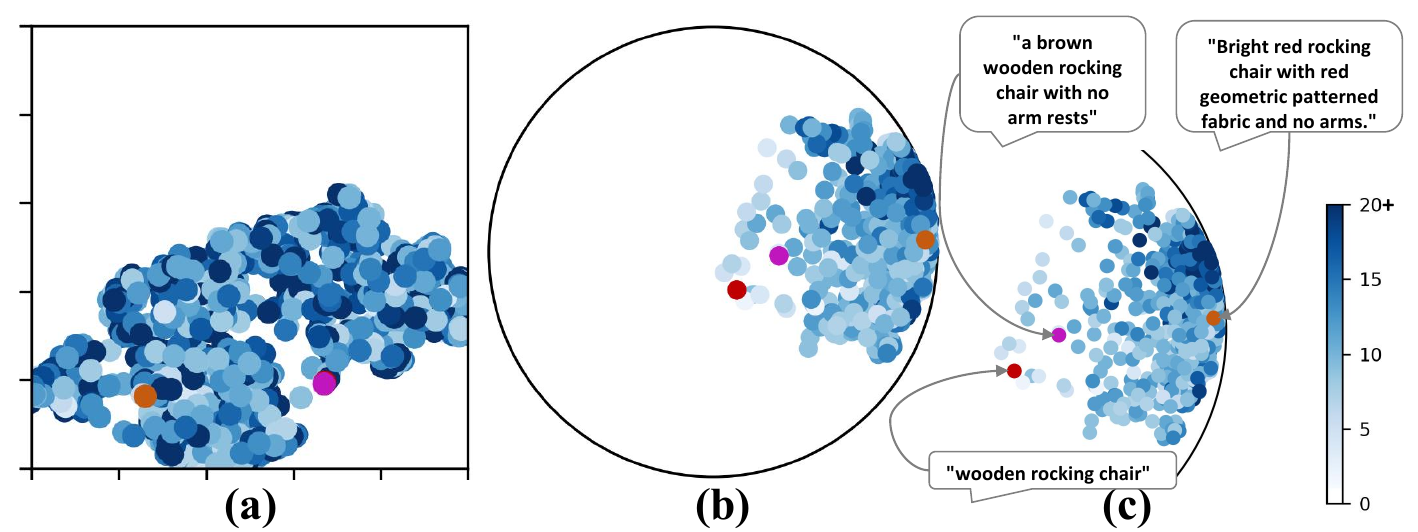}
    \caption{UMAP visualization of 2D text embeddings of 1000 random training samples. The color bar indicates the length of the text. (a): 2D text embeddings learned by SDFsuion~\cite{sdfusion2023} in Euclidean space. (b): 2D text embeddings learned by our method in hyperbolic space. (c) is the magnified view of (b).}
    \label{fig:text-hiera}
\end{figure}

\paragraph{The hierarchy of text feature} We visualize 2D text embeddings of 1000 random training samples in Figure~\ref{fig:text-hiera}. The dot color represents the length of the text. The light blue dot refers to the short text, that is general text without detailed information, like ``a chair". The dark blue refers to the long text, that is detailed text, like ``The silver and brown color iron chair with four legs and sponge.". As illustrated in Figure~\ref{fig:text-hiera} (a), the 2D text embeddings learned by SDFusion~\cite{sdfusion2023} are cluttered because the text encoder of SDFusion~\cite{sdfusion2023}, BERT in Euclidean space~\cite{bert}, cannot capture the text-shape hierarchy. In contrast, it can be observed from Figure~\ref{fig:text-hiera} (c) that the text length of text embeddings learned by our hyperbolic text-image encoder increases along the radius. It represents that features of general texts are close to the center point, and features of detailed text are near the boundary, exhibiting a hierarchical structure in hyperbolic space. Furthermore, we highlight a sample of text hierarchy in Figure~\ref{fig:text-hiera}, the red point refers to a general text, "wooden rocking chair", a pink point refers to a middle-level text, "a brown wooden rocking chair with no arm rests", and an orange point refers to a more detailed text, "Bright red rocking chair with red geometric patterned fabric and no arms.". It can be observed that the text embeddings of these points in Figure~\ref{fig:text-hiera} (a) do not follow the hierarchical structure, while the text embeddings of these points in Figure~\ref{fig:text-hiera} in Figure~\ref{fig:text-hiera} exhibits the hierarchical structure.

\begin{figure}[hp]
    \centering
    \includegraphics[width=1\linewidth]{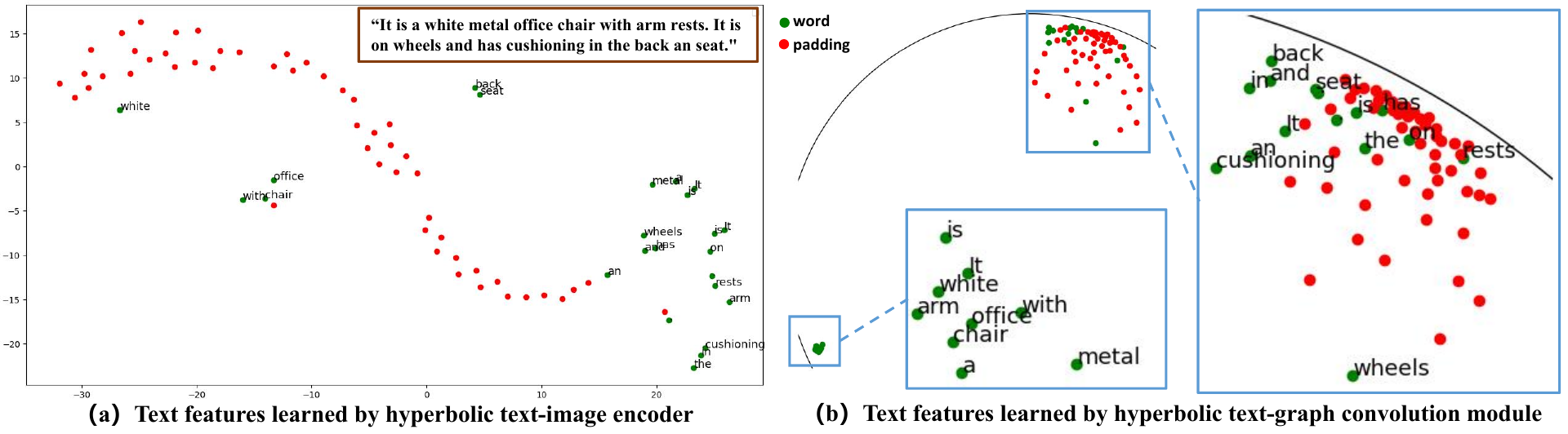}
    \caption{Text features learned by our HTIE and HTGC modules.}
    \label{fig: two_branch_vis}
\end{figure}

\paragraph{Learned two kinds of text features} Employing these two kinds of text features aims to leverage both the inherent sequential property and linguistic structures of text. Depicted in Figure~\ref{fig: two_branch_vis}, the feature distribution in Figure~\ref{fig: two_branch_vis}(\textbf{a}) showcases its sequential nature, while features in Figure~\ref{fig: two_branch_vis}(\textbf{b}) are more consistent with linguistic structure correlation.

\begin{figure}
    \centering
    \includegraphics[width=0.8\linewidth]{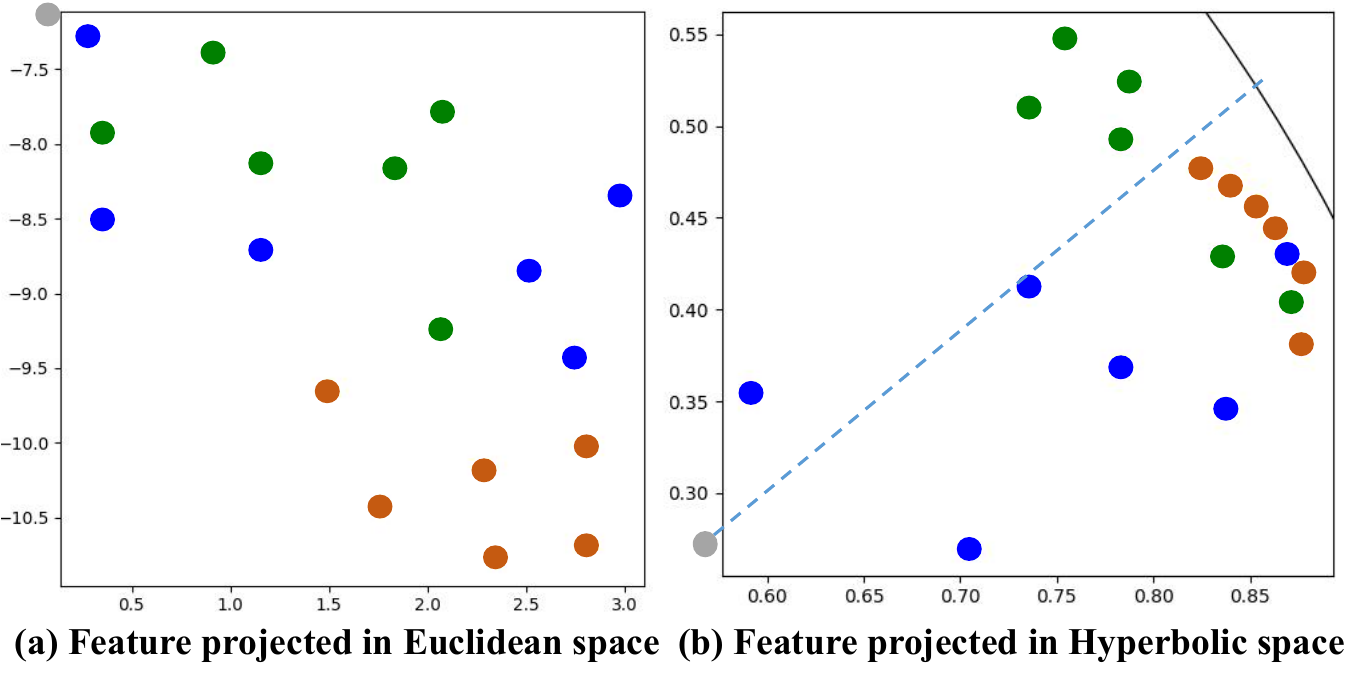}
    \caption{Features of 3D shape projected in the Euclidean space and hyperbolic space by Umap. The blue dot signifies deep features, green denotes middle features, and orange represents shallow features. The blue line is the radius of the \Poincare Ball.\vspace{-3mm}}
    \label{fig:3d-shape-hei}
\end{figure}

\paragraph{3D shape feature visualization.} We also illustrate the feature distribution in Euclidean and hyperbolic space, as shown in Figure~\ref{fig:3d-shape-hei}. It is observed that features in Euclidean space do not exhibit a tree-like hierarchical structure, conversely to those in hyperbolic space, which expand from the gray origin to the deep features in blue, the middle features in green, and finally to the shallow features in orange. It indicates the denoiser supervised by our hyperbolic hierarchical loss guarantees the tree-like hierarchical structure of 3D shape.

\begin{figure}[htb]
    \centering
    \includegraphics[width=1\linewidth]{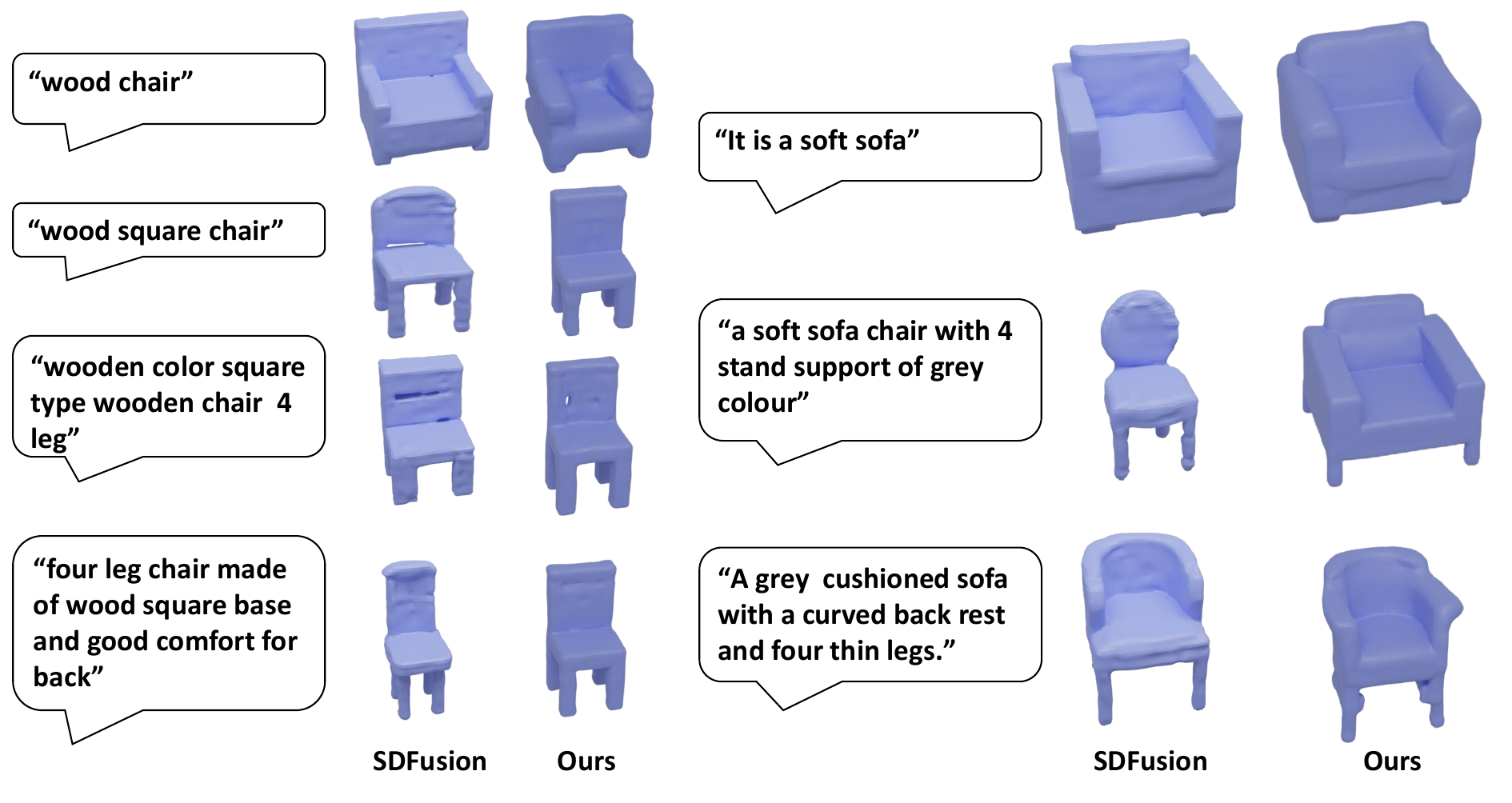}
    \caption{More visualizations for capturing text-shape hierarchy. }
    \label{fig: t2s1}
\end{figure}

\begin{figure}[htb]
    \centering
    \includegraphics[width=1\linewidth]{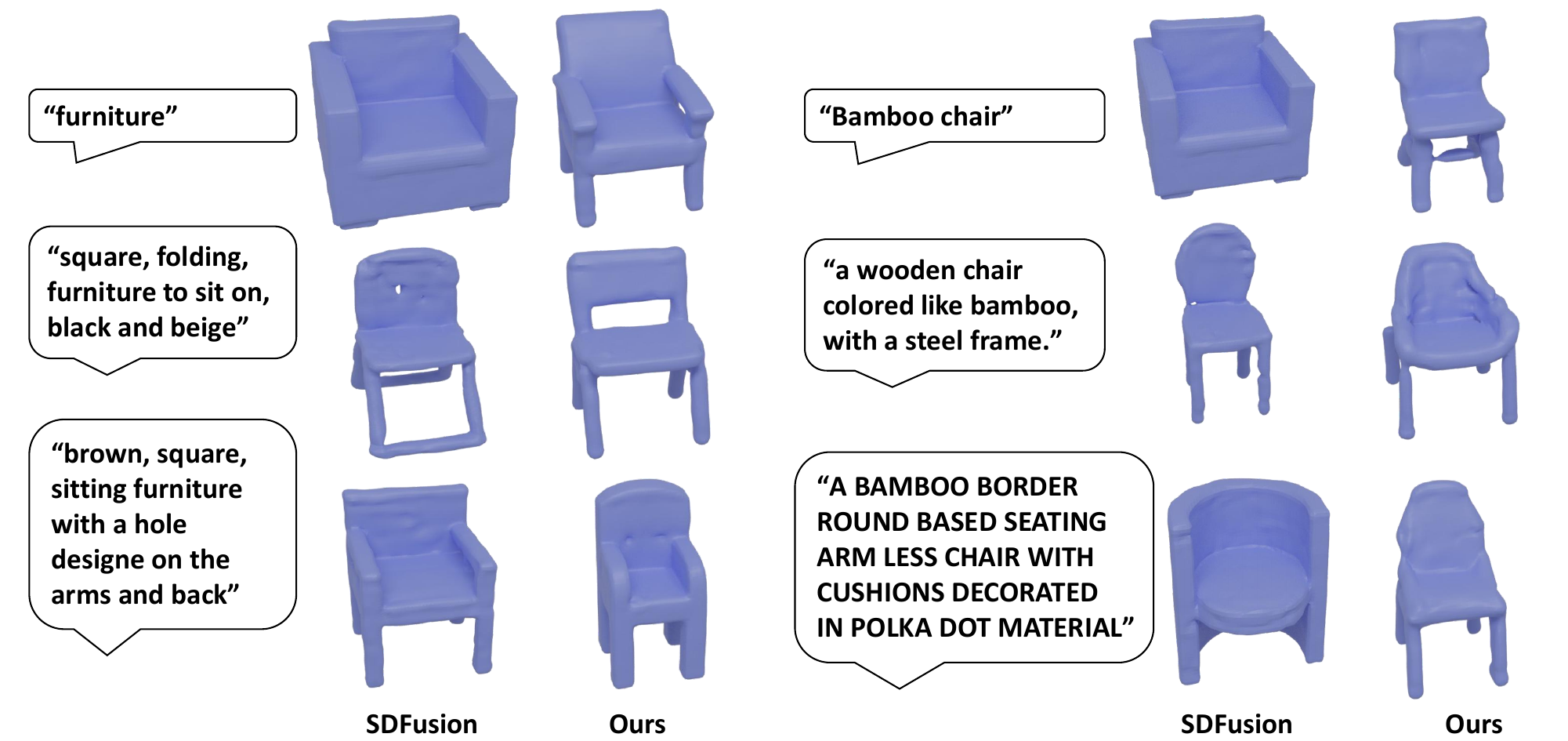}
    \caption{More visualizations for capturing text-shape hierarchy. }
    \label{fig: t2s2}
\end{figure}

\begin{figure}[htb]
    \centering
    \includegraphics[width=1\linewidth]{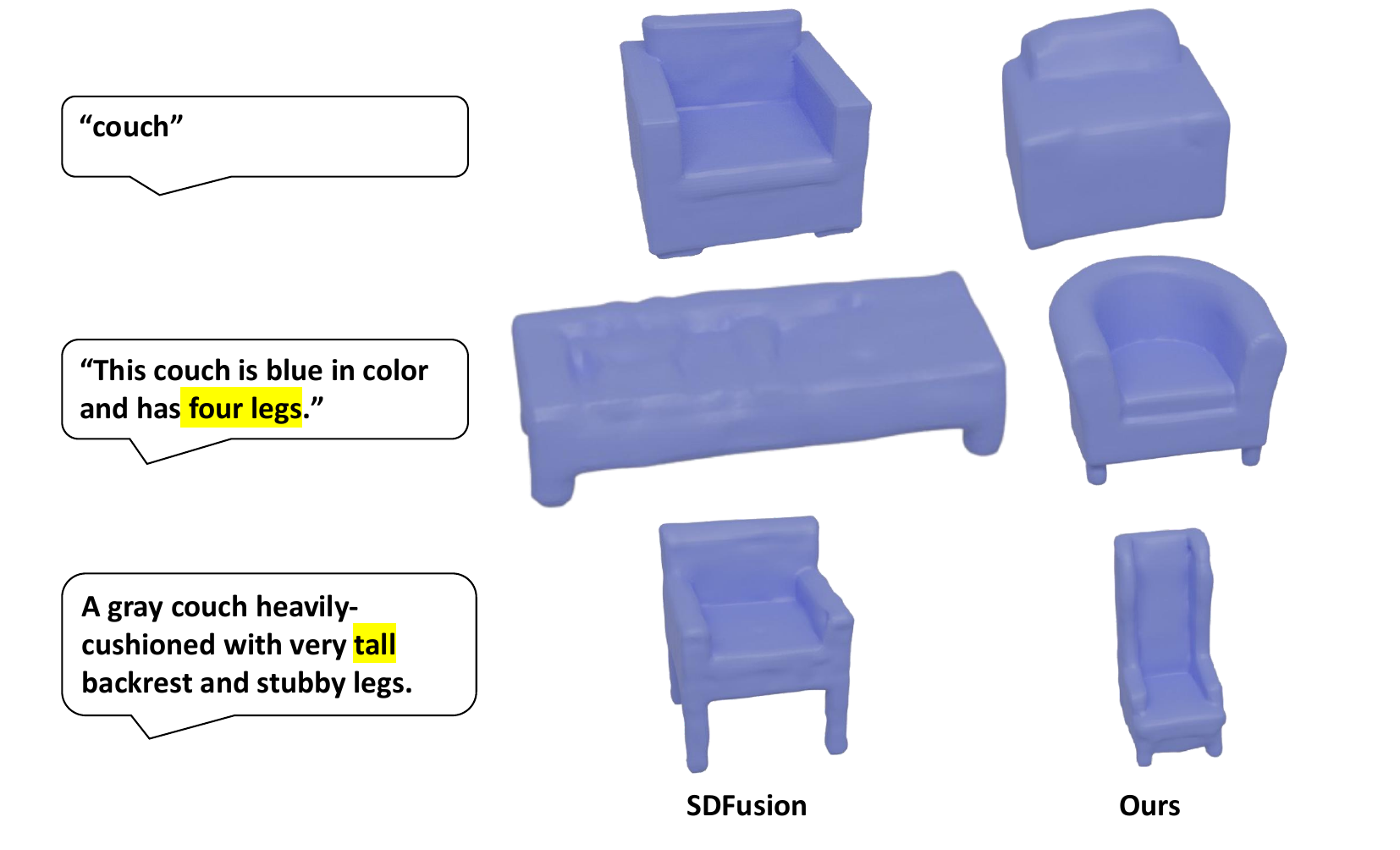}
    \caption{More visualizations for capturing text-shape hierarchy. }
    \label{fig: t2s3}
\end{figure}

\paragraph{More results and analysis for text-shape hierarchy.} In Table~\ref{tab: text-shape hei}, we have provided a sample of hierarchical text, and the HMD between the generated shapes. In Table~\ref{tab: more text-shape hei}, we enumerate more samples to demonstrate our performance of capturing text-shape hierarchy. Moreover, We also provide more visualizations for capturing text-shape hierarchy in Figure~\ref{fig: t2s1}, Figure~\ref{fig: t2s2}, and Figure~\ref{fig: t2s3}. It can be observed that 3D shapes generated by our method exhibit a hierarchy from general texts to detailed texts. In contrast, the shapes generated by SDFusion~\cite{sdfusion2023} from general texts to detailed texts do not correlate.

\begin{table}[tb]
\resizebox{\linewidth}{!}{
\begin{tabular}{lll}
\hline
\multicolumn{1}{c}{\multirow{2}{*}{Text}} & \multicolumn{2}{c}{HMD$\downarrow$ }                                \\
\multicolumn{1}{c}{}                      & \multicolumn{1}{c}{SDFusion~\cite{sdfusion2023}} & \multicolumn{1}{c}{Ours} \\
\hline
``wood chair"                                   &            -                  &              -           \\
``wood square chair"                    &       1.84            &     \textbf{0.53}          \\
``wooden color square type wooden chair  4 leg"    &        0.40      &      \textbf{0.04}          \\
``four leg chair made of wood square base and good comfort for back"     &      0.87    &     \textbf{0.03}      \\
\hline
``modern chair"                                   &            -                  &              -           \\
``Modern silver and gray office chair."                    &       2.57            &     \textbf{0.24}          \\
``Modern office chair with three legs made of metal and fibre made black seat."    &        1.18      &      \textbf{0.34}          \\
\hline
``It is a soft sofa"                                   &            -                  &              -           \\
``a soft sofa chair with 4 stand support of grey color"                    &       2.35            &     \textbf{0.29}          \\
``A grey cushioned sofa with a curved back rest and four thin legs."    &   \textbf{0.82}      &   0.93\\
\hline
``furniture"   & -  & - \\
``square, folding, furniture to sit on, black and beige"  & 1.82  & \textbf{0.32} \\
``brown, square, sitting furniture with a hole designe on the arms and back"  & 0.67 & \textbf{0.29} \\
\hline
``couch"  & -  &-  \\
``This couch is blue in color and has four legs."  & 2.76 & \textbf{0.48}\\
``A gray couch heavily-cushioned with very tall backrest and stubby legs." & \textbf{1.38} & 1.63\\
\hline
``Bamboo chair"  & - & - \\
``a wooden chair colored like bamboo, with a steel frame." & 2.24 & \textbf{0.45}\\
``A BAMBOO BORDER ROUND BASED SEATING ARM LESS \\ CHAIR WITH CUSHIONS DECORATED IN POLKA DOT MATERIAL" & 1.42 & \textbf{0.23} \\
\hline
\end{tabular}}
\caption{The result of comparing the performance of capturing the text-shape hierarchical structure. \vspace{-3mm}}
\label{tab: more text-shape hei}
\end{table}

\end{document}